# Generating Diverse and Meaningful Captions:
## Unsupervised Specificity Optimization for Image Captioning

Annika Lindh[1,2], Robert J. Ross[1,2], Abhijit Mahalunkar[2],
Giancarlo Salton[1,2] and John D. Kelleher[1,2]

[1] ADAPT Centre, Dublin, Ireland
[2] Dublin Institute of Technology (DIT), Dublin, Ireland
{annika.lindh,robert.ross,abhijit.mahalunkar,
giancarlo.salton,john.d.kelleher}@dit.ie

**Abstract.** Image Captioning is a task that requires models to acquire a multimodal understanding of the world and to express this understanding in natural language text. While the state-of-the-art for this task has rapidly improved in terms of n-gram metrics, these models tend to output the same generic captions for similar images. In this work, we address this limitation and train a model that generates more diverse and specific captions through an unsupervised training approach that incorporates a learning signal from an Image Retrieval model. We summarize previous results and improve the state-of-the-art on caption diversity and novelty. We make our source code publicly available online[1].

**Keywords:** Image Captioning, Diversity, Specificity, Computer Vision, Natural Language Generation, Natural Language Processing, Image Retrieval, Multimodal Training, Neural Networks, Deep Learning, Machine Learning, Contrastive Learning, MS COCO.

## 1 Introduction

Image Captioning is a task that requires models to acquire a multimodal understanding of the world and to express this understanding in natural language text, making it relevant to a variety of fields from human-machine interaction to data management. The practical goal is to automatically generate a natural language caption that describes the most relevant aspects of an image. Most state-of-the-art neural models are built on an encoder-decoder architecture where a Convolutional Neural Network (CNN) acts as the encoder for the image features that are fed to a Recurrent Neural Network (RNN) which generates a caption by acting as a decoder. It is also common to include one or more attention layers to focus the captions on the most salient parts of an image. The standard way of training is through Maximum Likelihood Estimation (MLE) by using a cross-entropy loss to replicate ground-truth human-written captions for corresponding images. Recent Image Captioning models of this kind [1, 11, 12, 28] have shown impressive results, much thanks to the powerful language modelling capabilities of Long

---

[1] https://github.com/AnnikaLindh/Diverse_and_Specific_Image_Captioning

Short-Term Memory (LSTM) [15] RNNs. However, although MLE training enables models to confidently generate captions that have a high likelihood in the training set, it limits their capacity to generate novel descriptions. Their output exhibits a disproportionate replication of common n-grams and full captions seen in the training set [9, 11, 26].

Contributing to this problem is a combination of biased datasets and insufficient quality metrics. While the main benchmarking dataset for Image Captioning, MS COCO, makes available over 120k images with 5 human-annotated captions each [6], the selection process for the images suggests a lack of diversity in both content and composition [11, 20]. Furthermore, the standard benchmarking metrics, based on n-gram level overlap between generated captions and ground-truth captions, reward models with a bias towards common n-grams. This leads to the (indirect and unwanted) consequence of incentivizing models that output generic captions that are likely to fit a range of similar images, despite missing the goal of describing the relevant aspects specific to each image.

In this paper, we propose a model that produces more *diverse and specific* captions by integrating a Natural Language Understanding (NLU) component in our training which optimizes the *specificity* of our Natural Language Generation (NLG) component. Our main contribution is an *unsupervised specificity-guided training approach* that improves the *diversity* and semantic accuracy of the generated captions. This approach can be applied to neural models of any multimodal NLG task (e.g. Image Captioning) where a corresponding NLU component can be made available.

We begin with an analysis of metrics for measuring caption quality in Section 2, where we define what we believe to be an informative set of metrics for our target. Following this, in Section 3 we describe our novel training approach along with the technical details of the NLG (our Image Captioning model) and NLU components for our experiments. In Section 4 we outline the experiments we undertook to evaluate our approach, followed by a discussion of our quantitative and qualitative results in Section 5. We review related work in Section 6 before presenting our conclusions and suggestions for future work in Section 7.

## 2 Measuring Caption Quality

The subjectivity in what defines a *good* caption, has made it difficult to identify a single metric for the overall quality of Image Captioning models [5, 26]. Benchmarking methods from Machine Translation [3, 19, 23] have been appropriated, while other somewhat similar methods such as CIDEr [27] have been proposed specifically for assessing the quality of image captions. All these approaches unfortunately have a strong focus on replicating common n-grams from the ground-truth captions [5] and do not take into account the richness and diversity of human expression [9, 26]. Moreover, it has been found that this class of metrics suffers from poor correlations with human evaluation, with CIDEr and METEOR having the highest correlations among them [5].

With the recognition of these limitations, there has been a growing interest in developing metrics that measure other desirable qualities in captions. SPICE [2] is a recent addition which measures the overlap of content by comparing automatically generated scene-graphs from the ground-truth and generated captions. While being a relevant addition, it does not solve the problem of generic captions. Rare occurrences and more detailed descriptions are more likely to incur a penalty than common concepts; e.g. correctly specifying *a purple flower* where the ground-truth text omits its color would register a false positive for the color. This, again, encourages the "safe" generic captions that we want to move away from.

### 2.1 Diversity Metrics

In an effort to measure the amount of generic captions produced by various Image Captioning models, [11] explores the concept of *caption diversity*. More recently, this concept has been employed as the focus for training and evaluation [26, 29], and it has been proposed that improving caption diversity leads to more human-like captions [26]. This research direction is still new and lacks clear benchmarks and standardized metrics. We propose the following set of metrics to evaluate the diversity of a model:

— *novelty* - percentage of generated captions where exact duplicates are not found in the training set [11, 26, 29]
— *diversity* - percentage of distinct captions (where duplicates count as a single distinct caption) out of the total number of generated captions [11]
— *vocabulary size* - number of unique words used in generated captions [26]

### 2.2 Meaningful Diversity Through Specificity

The diversity metrics alone do not tell us if a diverse model is more meaningful or if it simply introduced more noise. We argue that improving the *specificity* of the captions is essential to producing a meaningful increase in diversity. Our hypothesis is that by directly increasing the specificity, we will also achieve a higher diversity since diversity is a necessity for specificity. By improving both the specificity and diversity, we expect to generate qualitatively better captions that are less generic.

For this purpose, we propose a training architecture where a specificity loss is inferred by a separately trained Image Retrieval model. Specificity is measured by two standard Image Retrieval metrics:

— *recall at k* - percentage of generated captions resulting in the original image being found in the top *k* candidates retrieved by the Image Retrieval model
— *mean rank* - mean rank given by the Image Retriever to the correct image based on its generated caption

## 3 Optimizing for Specificity

To train a model that produces more diverse and meaningful captions, we propose to use an Image Retrieval model to improve the caption specificity of an Image Captioning model. In Image Retrieval tasks, a given query must be specific enough to retrieve the correct image among other, possibly similar, images. In this paper, we investigate whether the error signal from an Image Retrieval model can improve caption specificity in an Image Captioning model, and whether these more specific captions are also more diverse.

The training process is inspired by [22] where the task is to generate Referring Expressions that unambiguously refer to a region of an image; their solution is to introduce a Region Discriminator that measures the quality of their generated expressions. Their method is in turn inspired by Generative Adversarial Networks (GANs) in which a Generator and a Discriminator are in constant competition - the Discriminator aims to distinguish between real and generated data, while the Generator aims to generate data that the Discriminator cannot tell apart from the real data [13]. In [22], the training is cooperative rather than competitive; both systems adjust to the other to provide the best joint results.

We take a slightly different approach from both the joint training in [22] and recent applications of GAN training in Image Captioning [9, 26]. Instead of allowing both systems to learn from each other, we freeze the NLU side and allow only the NLG to learn from the NLU; the NLU model is pre-trained on ground-truth captions, without any input from the NLG. Consequently, we avoid one of the problems observed in [22] where both systems adapt to each other and develop their own protocol of communication which gradually degrades the resemblance to human language. We also avoid the instability in training and difficulty in loss monitoring commonly seen in GANs.

### 3.1 Model Architecture

To demonstrate our training approach, we practically apply it to a neural Image Captioning model proposed in [1] which uses an encoder-decoder architecture with region-based attention. For our experiments, we use a publicly available re-implementation [21]. To leverage the fluency gained from MLE training, the model is pre-trained to minimize the cross-entropy loss $L_{XE}$ for each ground truth sequence $y_{1:T}$ when conditioned on an image $I$ and the attended image features $i_{1:T}$:

$$L_{XE}(\theta) = -\sum_{t=1}^{T} log\big(p\theta(y_t|y_{1:t-1}, i_t, I)\big). \tag{1}$$

The pre-trained model also provides a strong baseline to compare to. The model architecture, illustrated in **Fig. 1**, consists of a ResNet-101 [14] CNN pre-trained on the ImageNet [25] dataset, followed by an LSTM for attention modelling, and a second LSTM that generates the captions. (Unlike [1], the attention-regions are 14x14 regions over the final convolutional layer instead of using a region proposal network.) During our specificity training, the CNN layers remain frozen while we update the weights of the two LSTMs.

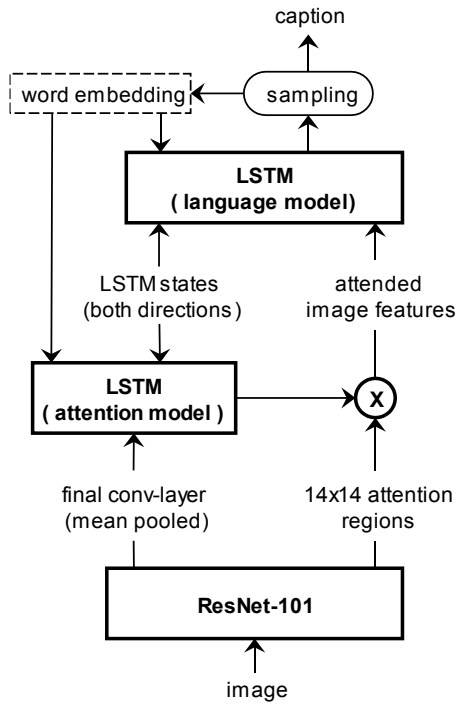
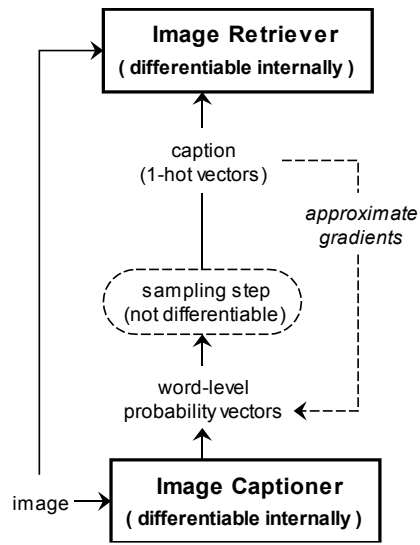

**Fig. 1.** Our Image Captioning model architecture.

**Fig. 2.** Interactions between the Image Captioning and Image Retrieval models during training.

For our NLU component, we use the neural Image Retrieval model from the SentEval toolkit [8]; the NLU is pre-trained on ground-truth data and remains frozen during our specificity training. Given an image-caption pair, it produces the loss and gradients for our Image Captioning model by projecting the image and caption into the same space to estimate their similarity. The image embeddings are acquired by a ResNet-101 trained on ImageNet, and the captions are embedded using InferSent [7] with GloVe [24] word embeddings.

### 3.2 Specificity Loss Functions

We define four different loss functions to be calculated by our NLU component, each used in one of the model variations. The first two improve the individual similarity of a caption to its corresponding image, while the latter two implement contrastive pair-wise versions of the first two.

Let $c$ be the projected caption embedding and let $i$ be the projected image embedding, both acquired by passing the generated caption $C$ and its original corresponding image $I_o$ through the Image Retrieval model. For the contrastive loss functions, let $I_c$ be a contrastive image chosen at random from the top 1% most similar images to $I_o$ based on its activations from the final convolutional layer of the encoder CNN. We can now define the dot product similarity loss $L_{DP}$, the cosine similarity loss $L_{Cos}$, the contrastive dot product loss $L_{CDP}$ and the contrastive cosine loss $L_{CCos}$. Equations 2 - 5 define the loss functions in terms of a single example; the final loss is the mean loss over all examples.

$$L_{DP}(C, I_o) = -(c \cdot i),  \qquad (2)$$

$$L_{Cos}(C, I_o) = -\frac{c \cdot i}{||c|| \cdot ||i||},  \qquad (3)$$

$$L_{CDP}(C, I_o, I_c) = max(0, c \cdot i_c - c \cdot i_o),  \qquad (4)$$

$$L_{CCos}(C, I_o, I_c) = max\left(0, \frac{c \cdot i_c}{||c|| \cdot ||i_c||} - \frac{c \cdot i_o}{||c|| \cdot ||i_o||}\right).  \qquad (5)$$

### 3.3 Training

The interactions between the NLU and NLG components are illustrated in **Fig. 2**. At each iteration, the Image Captioning model generates a full caption for a given image (or a set of captions for a batch of images). This involves a non-differentiable sampling step to convert the word-level probabilities into a sequence of discrete words represented by 1-hot encoded vectors. The caption is then fed to the Image Retriever along with its corresponding image, where both are passed through the embedding and projection steps.

The Image Retriever calculates one of the specificity losses defined in Section 3.2. To minimize this loss, we need to backpropagate the gradients through the Image Retrieval model's (frozen) layers and then back through the Image Captioning model's layers that we wish to update. This is not trivial since our forward pass includes a non-differentiable sampling step. To overcome this, we apply the Straight-Through method [4] and use the gradients with respect to the 1-hot encoding as an approximation for the gradients with respect to the probabilities before sampling. We empirically validate this approach by observing that our loss decreases smoothly. We also experimented with the similar Gumbel Straight-Through method [16] but observed no empirical benefit.

## 4 Experiment Design

All experiments are conducted in PyTorch[2]. Our implementation extends the code of the baseline Image Captioning model by replacing the MLE training with our specificity training. The Image Retrieval code is modified to calculate our specificity losses defined in Section 3.2. We use the Adam [18] optimizer with an initial learning rate of $1 \times 10^{-6}$ for the contrastive models and $1 \times 10^{-7}$ for the other two models. Early stopping is used based on the lowest mean rank on the validation set. The contrastive models trained for about 190k iterations on the randomly shuffled training set, while the non-contrastive models trained for about 250k iterations, all using a batch-size of 2. When sampling from the final models on the test set, any tokens that are duplicates of the immediately previous token are automatically removed since such duplicates were an issue in our non-contrastive models; we do the same for all our models, including the baseline, for a fair comparison.

---

[2] https://pytorch.org/

### 4.1 Dataset

We use the MS COCO dataset [20] with the Karpathy 5k splits [17], containing 113k images for training and 5k each for validation and test, with 5 captions for each image. The same splits were used for both the NLG and the NLU, including pre-training, ensuring that we have no overlap between training, validation and test data and that our improvements do not come from bridging a gap between different datasets. Note that the specificity training does not require any extra data in addition to that used during pre-training. Furthermore, since the labels are not used during our specificity training, one could also make use of unlabeled data. All splits were pre-processed by lower-casing all words and removing punctuation. Any words appearing less than 5 times in the training set were replaced by the UNK token, resulting in a vocabulary size of 9487 (including the UNK token).

**Table 1.** Diversity and Specificity. Our models are named after the loss functions defined in 3.2. All metrics are percentages except Vocab Size and Mean Rank which are absolute numbers. Higher is better except for Mean Rank where lower is better. Results for *a* as reported in [11].

| | Diversity and Specificity | | | | | | |
|---|---|---|---|---|---|---|---|
| | Diversity | Novelty | Vocab Size | R@1 | R@5 | R@10 | Mean Rank |
| D-ME+DMSM [12][a] | 47.0[a] | 70.0[a] | - | | | | |
| Adv-samp [26] | - | 73.9 | **1616** | | | | |
| DP (Ours) | 79.12 | 76.66 | 1029 | 10.38 | 31.38 | 44.48 | 33.70 |
| Cos (Ours) | 79.16 | 76.66 | 1034 | 10.04 | 30.66 | 43.54 | 35.25 |
| CDP (Ours) | **84.48** | **77.49** | 1064 | **12.80** | 36.16 | 49.67 | 32.79 |
| CCos (Ours) | 84.37 | 77.29 | 1052 | 12.53 | **36.19** | **50.00** | **32.30** |
| Baseline | 76.26 | 69.08 | 812 | 10.82 | 30.42 | 43.32 | 39.25 |

## 5 Results and Discussion

The models we compare to are the best models in terms of diversity from [11, 26], using the single best caption after re-ranking for the latter. We also report the specificity metrics used for our training goals. The results for specificity would not be directly comparable to models using other external systems, but they are relevant when assessing our own models and verifying that our increase in diversity follows from an increase in specificity. Results from our contrastive models are averaged over 3 runs each. The non-contrastive models are based on single runs.

As can be seen in **Table 1**, our models demonstrate increased diversity and novelty, outperforming previously reported results. The vocabulary size also increases but is

48lower than in [26]. When it comes to the specificity metrics, our contrastive models have the advantage over our non-contrastive ones. They all improve the overall mean rank, but the latter do not show the increase in smaller $k$ recalls that the contrastive models do. This is not surprising since the contrastive models specifically minimize their loss in comparison to similar images, while the non-contrastive ones increase their semantic similarity in isolation. The higher specificity of the contrastive models is also accompanied by higher values in diversity and novelty.

**Table 2.** Novelty and diversity per image with up to 10 candidates; novelty and diversity was not reported for the single-best-caption output.

| Diversity metrics for multi-candidate models | | |
|---|---|---|
| | Diversity within candidates | Novelty within candidates |
| CVAE [29] | 11.8 | **82.0** |
| GMM-CVAE [29] | 59.4 | 80.9 |
| AG-CVAE [29] | **76.4** | 79.5 |

For completeness, we include the best models from [29] in **Table 2**; however, they only report diversity results on multiple (up to 10) candidates per image (where duplicates of a novel caption are counted as multiple novel captions), so they are not directly comparable to the single-best-caption models. Note that [12, 29] use different data splits, while our models and [26] use the Karpathy 5k splits [17].

**Table 3.** Standard text metric results for single-best-caption models. All metrics are n-gram based except for SPICE which is based on scene graphs automatically inferred from the captions.

| Standard text metrics B-$n$ = BLEU-$n$   R-L = ROUGE-L   M = METEOR   C = CIDEr   S = SPICE | | | | | | | | |
|---|---|---|---|---|---|---|---|---|
| | B-1 | B-2 | B-3 | B-4 | R-L | M | C | S |
| D-ME+DMSM [12] | - | - | - | 0.257 | - | 0.236 | - | - |
| Adv-samp [26] | - | - | - | - | - | 0.236 | - | 0.166 |
| CVAE [29] | 0.698 | 0.521 | 0.372 | 0.265 | 0.506 | 0.225 | 0.834 | 0.158 |
| GMM-CVAE [29] | 0.718 | 0.538 | 0.388 | 0.278 | 0.516 | 0.238 | 0.932 | 0.170 |
| AG-CVAE [29] | 0.716 | 0.537 | 0.391 | 0.286 | 0.517 | 0.239 | 0.953 | 0.172 |
| DP (Ours) | 0.725 | 0.556 | 0.409 | 0.297 | 0.527 | 0.247 | 0.953 | 0.184 |
| Cos (Ours) | 0.725 | 0.556 | 0.409 | 0.297 | 0.527 | 0.247 | 0.953 | 0.184 |
| CDP (Ours) | 0.736 | 0.564 | 0.417 | 0.306 | 0.533 | 0.251 | 0.977 | 0.188 |
| CCos (Ours) | 0.737 | 0.565 | 0.419 | 0.307 | 0.533 | 0.253 | 0.980 | 0.188 |
| Baseline | **0.746** | **0.579** | **0.432** | **0.320** | **0.545** | **0.262** | **1.036** | **0.197** |

In **Table 3**, we report results on the standard text metrics. As expected, we see a slight decrease in these metrics when moving away from safer generic captions. They are, however, still in line with our state-of-the-art baseline and slightly stronger than previous diversity-focused models.

### 5.1 Qualitative Analysis

Our contrastive models tend to generate more specific (and accurate) captions while the baseline model prefers common patterns from the training data, as can be seen in the leftmost images in **Fig. 3**. As is particularly evident in the bottom image, our contrastive models pay more attention to the image content (i.e. mentioning the dog) while the baseline model pays more attention to the language priors (i.e. assuming the presence of a surfboard on the beach). The rightmost image shows a failure case where our contrastive models focus on the wooden structure (which is more unique in this context) while omitting the skateboard (which is more common, but also more relevant).

The improvement in diversity and specificity is not achieved by simply producing longer captions; the average caption length for the baseline, contrastive and non-contrastive models were 9.6, 9.4 and 8.9 words respectively.

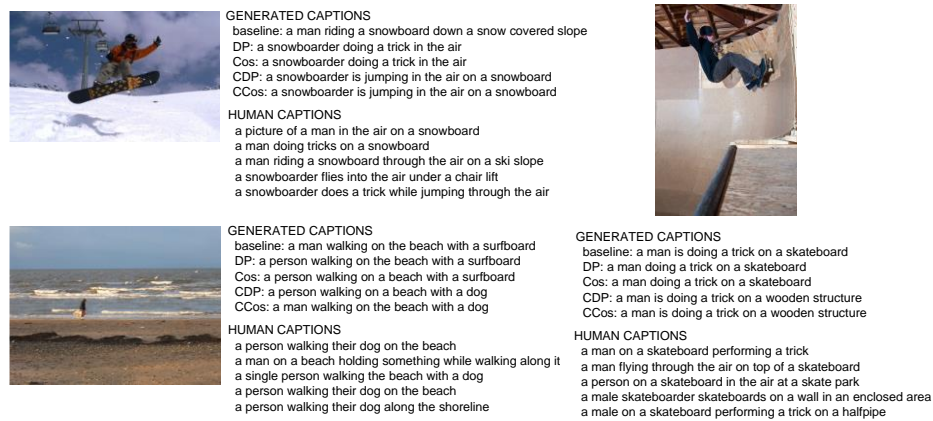

**Fig. 3.** Examples of generated captions and human annotations. The rightmost image shows a failure case where specificity took precedence over relevance.

## 6 Related Work

While Image Captioning has received a lot of attention, the focus has mainly been on n-gram metric results. [11] provides some insight into the problems that follow from the standard training and metrics, noting the lack of *diversity* observed in captions from state-of-the-art neural models. More recently, this has led to some initial attempts at improving caption diversity.

In [9], a GAN model conditioned on the image is proposed. The authors do not report any quantitative results for diversity, but they show qualitative examples after manually adjusting the variance of the input to the GAN. This demonstrates the ability of LSTMs to produce fluent captions under noisy conditions, leading to some variation in the output. We observed a similar effect in experiments with noise-based gradients. However, such methods are not constrained to produce *meaningful diversity* (as discussed in Section 3) and the level of noise that is appropriate for one caption might be too high for another.

Another example of GAN training is [26] where the Discriminator classifies whether a multi-sample set of captions are human-written or generated. In contrast, our evaluator only requires a single caption and uses a much simpler loss function. Furthermore, we let the NLU remain frozen during training, making the training stable and producing more informative learning curves.

A similar approach can be found in [10] where Contrastive Learning is used in a GAN-like setting. In contrast to our approach which is unsupervised after pre-training, theirs require image-caption pairs both during and after pre-training. Similar to our work, they are motivated by a specificity goal; unfortunately, they do not report results on any diversity metrics.

## 7    Conclusion

With this work, we have highlighted an important limitation in current Image Captioning research. We provided a discussion on the limitations of current evaluation metrics and proposed a set of metrics related to *diversity* while emphasizing the importance of *meaningful* diversity. Our work summarizes previously reported results and contributes a new state-of-the-art in this area in terms of diversity and novelty. The code for our model and training approach is made publicly available online to encourage further research.

To conclude, we believe that the standard MLE training has both benefits and drawbacks for Image Captioning and that much can be gained by combining it with additional optimization terms. By including an Image Retrieval learning signal, we introduced an additional dimension to our model's training by including text-to-image understanding in addition to its original image-to-text target.

We suggest further research into training approaches that incentivizes multimodal models to build a more complete, bi-directional understanding of its modalities. Additionally, we encourage further exploration of evaluation methods that assess additional desirable qualities in automatically generated captions.

**Acknowledgments.** This research was supported by the ADAPT Centre for Digital Content Technology which is funded under the SFI Research Centres Programme (Grant 13/RC/2106) and is co-funded under the European Regional Development Fund.